%% file: aaai24.tex
\documentclass[letterpaper]{article} % DO NOT CHANGE THIS
\usepackage{aaai24}  % DO NOT CHANGE THIS
\usepackage{times}  % DO NOT CHANGE THIS
\usepackage{helvet}  % DO NOT CHANGE THIS
\usepackage{courier}  % DO NOT CHANGE THIS
\usepackage[hyphens]{url}  % DO NOT CHANGE THIS
\usepackage{graphicx} % DO NOT CHANGE THIS
\usepackage{natbib}  % DO NOT CHANGE THIS AND DO NOT ADD ANY OPTIONS TO IT
\usepackage{caption} % DO NOT CHANGE THIS AND DO NOT ADD ANY OPTIONS TO IT
\usepackage{algorithm}
\usepackage{algorithmic}

\usepackage{booktabs}
\usepackage{amsmath}
\usepackage{amssymb}
\usepackage{pifont}
\usepackage{fontawesome}
\usepackage{multirow}
\usepackage{endnotes}
\usepackage[table,xcdraw]{xcolor}
\usepackage{tcolorbox}
\usepackage{makecell}
\usepackage{balance}    % balance space

\usepackage[pagebackref]{hyperref} % return ref page
\usepackage{hyperref}
\hypersetup{hidelinks} % hide box
\definecolor{lightblue}{rgb}{0, 0, 0.8}
\definecolor{lightgreen}{rgb}{0, 0.8, 0.0}
\hypersetup{colorlinks=true, citecolor=lightblue, linkcolor=red, urlcolor=lightblue}

\usepackage{newfloat}
\usepackage{listings}
\DeclareCaptionStyle{ruled}{labelfont=normalfont,labelsep=colon,strut=off} % DO NOT CHANGE THIS
\floatstyle{ruled}
\newfloat{listing}{tb}{lst}{}
\floatname{listing}{Listing}
\nocopyright %-- Your paper will not be published if you use this command
\title{Language-Assisted 3D Scene Understanding}
\author{
    Yanmin Wu\textsuperscript{\rm 1},
    Qiankun Gao\textsuperscript{\rm 1},
    Renrui Zhang\textsuperscript{\rm 2},
    Jian Zhang\textsuperscript{\rm 1}\thanks{ \textit{Corresponding author}.}
}
\affiliations{
    \textsuperscript{\rm 1} Shenzhen Graduate School, Peking University, China \quad
    \textsuperscript{\rm 2} The Chinese University of Hong Kong, China \\
    {\tt\small \{wuyanmin, gqk\}@stu.pku.edu.cn} \quad 
    {\tt\small zhangrenrui@link.cuhk.edu.hk} \quad 
    {\tt\small zhangjian.sz@pku.edu.cn}
}

\usepackage{bibentry}
\begin{document}

\maketitle

% show page number
\pagenumbering{arabic}
\thispagestyle{plain}

\begin{abstract}
The scale and quality of point cloud datasets constrain the advancement of point cloud learning. Recently, with the development of multi-modal learning, the incorporation of domain-agnostic prior knowledge from other modalities, such as images and text, to assist in point cloud feature learning has been considered a promising avenue.
Existing methods have demonstrated the effectiveness of multi-modal contrastive training and feature distillation on point clouds. However, challenges remain, including the requirement for paired triplet data, redundancy and ambiguity in supervised features, and the disruption of the original priors.
In this paper, we propose a \textbf{l}anguage-\textbf{as}sis\textbf{t}ed approach to \textbf{p}oint \textbf{c}loud feature \textbf{l}earning (\textbf{LAST-PCL}), enriching semantic concepts through LLMs-based text enrichment. We achieve de-redundancy and feature dimensionality reduction without compromising textual priors by statistical-based and training-free significant feature selection. Furthermore, we also delve into an in-depth analysis of the impact of text contrastive training on the point cloud.
Extensive experiments validate that the proposed method learns semantically meaningful point cloud features and achieves state-of-the-art or comparable performance in 3D semantic segmentation, 3D object detection, and 3D scene classification tasks.
\end{abstract}

%%%%%%%%%%%%%%%%%%%%%%%%%%%%%%%%%%%%%%%%%%%%%%%%%%%%%%%
% 1. Introduction
%%%%%%%%%%%%%%%%%%%%%%%%%%%%%%%%%%%%%%%%%%%%%%%%%%%%%%%
\section{Introduction}
3D point cloud understanding includes fundamental 3D visual tasks such as 3D semantic segmentation, 3D object detection, and 3D point cloud classification, critical for applications like robotics, autonomous driving, and mixed reality.
However, the challenges of data acquisition difficulty, high annotation costs, and small data scale have rendered 3D point cloud understanding a daunting task. Current research focuses on three aspects:
\textbf{1)} Exploring more efficient network architectures, such as various point cloud Transformers with novel attention mechanisms~\cite{wu2022point,yang2023swin3d}.
\textbf{2)} Utilizing self-supervised pretraining like mask-based reconstruction~\cite{yu2022point,zhang2022point} to discover intrinsic data representations.
\textbf{3)} Leveraging multimodal pre-trained models to enhance point cloud feature learning, distilling cross-domain semantic priors from sources like text or images~\cite{zhang2023learning,huang2023joint}.
In this work, we specifically aim to harness text features to augment point cloud understanding.

% motivation 1：text enrichment
Recent advancements in visual-language contrastive learning~\cite{radford2021learning} and multimodal large language models (MLLMs)~\cite{bao2022vlmo,li2023blip} show that using open-ended textual data for contrastive training improves visual data representation ability. For point cloud processing tasks, the promise of acquiring supervised knowledge from abundant domain-agnostic multimodal data is significant. The most intuitive approach~\cite{hegde2023clip,xue2023ulip} is to construct point-image-text triplets, where the pre-trained image encoder and text encoder supervise the training of the point cloud backbone. However, acquiring paired triplet data is not straightforward. It often requires rendering point clouds into images or establishing point-to-image associations based on camera poses, which is time-intensive and restricts scalability.
Although LG~\cite{rozenberszki2022language} explored training point clouds using text contrastive, the textual content comprised only categories with limited semantics.
Therefore, inspired by previous prompting works~\cite{guo2023viewrefer,zhang2023prompt}, we introduce using large language models (LLMs)~\cite{brown2020language} to create free-form, fine-grained descriptions for the categories of point clouds. This \textbf{text enrichment} not only enables point clouds to learn richer semantic concepts but also allows for more diverse textual queries, extending beyond the annotated categories.

% motivation 2: feature selection
Contrastive training demands that the dimensions of point cloud features align with those of image/text features. However, point cloud features are typically low-dimensional in the final network layer (\textit{e.g.}, 48, 64), while pre-trained encoders often extract high-dimensional image/text features (512, 768, or even 1024). Two solutions exist for feature alignment:
One approach~\cite{rozenberszki2022language, peng2023openscene} maintains the original high-dimensional image/text features and projects the point cloud features to the high-dimensional space. 
However, two issues arise: 1) Dense point cloud segmentation involves samples with over 10,000 points, making the per-point elevation to the higher dimension computationally expensive. 2) Original high-dimensional features might contain redundant or interfering information.
Another choice~\cite{jain2022bottom, wu2023eda} trains a projection layer to downscale image/text features to match point cloud dimensions, reducing computational costs. However, the projection layer disrupts the original textual priors, shifting text embeddings to a feature space fitted to the current training set and impairing generalization.
To address this, we propose a \textbf{statistical-based significant feature selection} method that reduces feature dimensionality and de-redundancy without disrupting the original textual priors, motivated by APE~\cite{zhu2023not}. Specifically, We compute intra-class similarity for distinct descriptions of the same category and inter-class variance for different categories. This generates a ranking, allowing us to reduce dimensions by selecting the top-$d$ channels that are most discriminative for inter-class recognition.

% contributions
The main contributions of this paper are as
follows:
\textbf{1)} We present three observations (Sec.~\ref{sec:Theoretical_Analysis}) regarding the influence of text-contrastive training on point cloud learning and empirically validate them (Sec.~\ref{sec:analysis}). We aspire for these insights to stimulate thoughtful considerations in the field.
\textbf{2)} We propose an efficient method for \textbf{l}anguage-\textbf{as}sis\textbf{t}ed \textbf{p}oint \textbf{c}loud feature \textbf{l}earning (\textbf{LAST-PCL}). In this method, we enhance semantic concepts by text enrichment (Sec. ~\ref{sec:text}) based on LLMs and achieve feature dimensionality reduction through statistical-based significant feature selection (Sec.~\ref{sec:selection}), which avoids redundancy and preserves text priors.
\textbf{3)} Extensive experiments (Sec.~\ref{sec:exp}) demonstrate that the LAST-PCL achieves SOTA or comparable performance across different tasks, including 3D semantic segmentation, 3D object detection, and 3D scene classification.

%%%%%%%%%%%%%%%%%%%%%%%%%%%%%%%%%%%%%%%%%%%%%%%%%%%%%%%
% 2. Related Work
%%%%%%%%%%%%%%%%%%%%%%%%%%%%%%%%%%%%%%%%%%%%%%%%%%%%%%%
\section{Related Work}

Due to the limited scale and noisy nature of point clouds, learning domain-agnostic knowledge from other modalities presents a promising avenue. Existing multi-modal point cloud feature learning can be categorized into two fashions. 

\textbf{(1) Multi-modal data shares a common encoder.} 
\textbf{\romannumeral1)} The most intuitive methods~\cite{wang2022p2p,zhang2022pointclip,zhu2022pointclip} render point clouds into images and extract features using an image encoder, achieving excellent zero-shot capabilities but constrained by the quality of rendered images and the scale of point clouds. 
\textbf{\romannumeral2)} 
Some studies employ frozen encoders from other modalities as the point cloud backbone. EPCL~\cite{huang2022frozen} and ACT~\cite{dong2023act} adopt frozen foundation Transformers, training only the point tokenizer and prompt components, inheriting the robust priors of foundation models with reduced training costs, but often leading to parameter redundancy. 
\textbf{\romannumeral3)} 
Joint-MAE~\cite{guo2023joint} and PCExpert~\cite{kang2023point} train multi-modal backbone from scratch, treating point clouds as specialized images and co-training with images to implicitly learn cross-modal representations, but it requires double the amount of data and computational resources.
\textbf{(2) Multi-modal feature distillation/contrastive training.} 
\textbf{\romannumeral1)} 
One option uses \textbf{fine-grained pixel-point cloud pairs}. I2P-MAE~\cite{zhang2023learning}, OpenScene~\cite{peng2023openscene}, UP-VL~\cite{najibi2023unsupervised}, and CLIP-FO3D~\cite{zhang2023clip} utilize pre-trained image encoders as teacher models to extract pixel features, supervising the training of point cloud student models through pixel-point relationships for open-vocabulary scene understanding. However, these methods involve time-consuming projection and complex multi-scale image feature extraction. 
\textbf{\romannumeral2)} 
Another option involves \textbf{object-level image-point-text triplets}. ULIP~\cite{xue2023ulip}, ReCon~\cite{qi2023contrast}, CG3D~\cite{hegde2023clip}, and Point-Bind~\cite{guo2023point} extract multi-modal features from pre-trained image/text encoders and perform contrastive training for point cloud backbones. Object-level datasets are typically needed for triplet pre-training. 
\textbf{\romannumeral3)} We focus on the task of \textbf{text-assisted scene-level point cloud understanding}. Despite some efforts~\cite{rozenberszki2022language, jain2022bottom, zeng2023clip2, huang2023joint} made in this area from the perspective of joint representation,
however, supervising with high-dimensional image/text features is irrational due to redundancy, ambiguity, and increased computational costs for point cloud projection. Additionally, reducing image/text features through projection would compromise textual priors and limit the model's generalization.
In contrast, our method sidesteps paired triplets and focuses on text-supervised point cloud learning. The proposed text enrichment boosts semantic concepts, while significant feature selection reduces redundancy and dimensions without compromising textual priors.

%%%%%%%%%%%%%%%%%%%%%%%%%%%%%%
% Method
%%%%%%%%%%%%%%%%%%%%%%%%%%%%%%
\begin{figure*}[!t]
\centering
\setlength{\abovecaptionskip}{2pt}
\includegraphics[width=1.0 \textwidth]{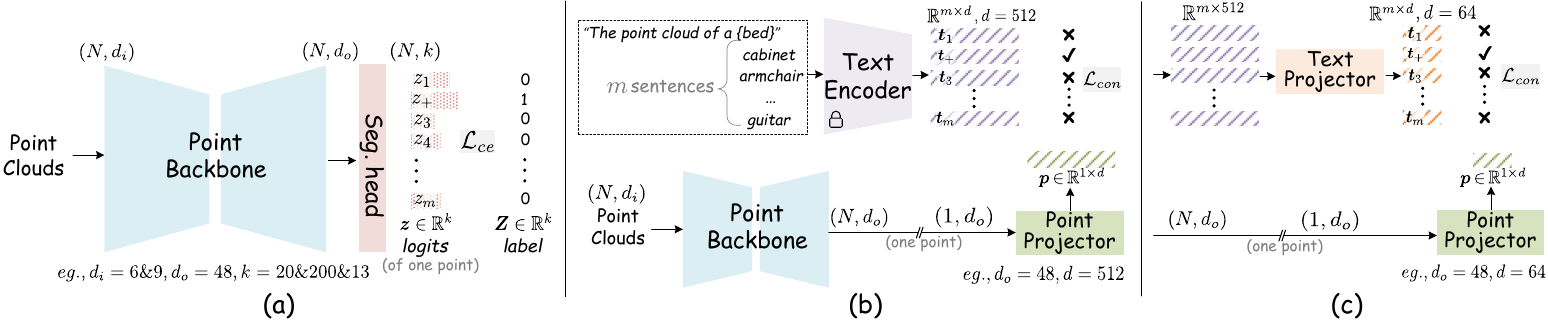} 
\caption{Different semantic segmentation pipelines. (a) Standard one-hot label supervision method. $N$ is the number of input points, $d_i$ is the input dimension, $d_o$ is the output dimension, and $k$ is the number of classes. (b) Text contrastive training, where text features keep the original dimension and point features are transformed to high dimension by a projection layer. $m$ is the number of sentences. (c) Text contrastive training, with text features transformed to low dimension via a projection layer.
}
\label{fig:three_type}
%\vspace{-10pt}
\end{figure*}

\section{Method}
We first present three observations in Sec.~\ref{sec:Theoretical_Analysis} regarding the impact of text-contrastive training on point cloud learning. These observations inspire our method, which is subsequently introduced in Sec.~\ref{sec:overview}-~\ref{sec:training_inference}.

\subsection{Can Text-Contrastive Training Help 3D Point Cloud Learning?}
\label{sec:Theoretical_Analysis}

\textbf{Formulation:} 
There are two training paradigms for the 3D cloud segmentation task. The standard approach uses one-hot labels and cross-entropy loss, as shown in Fig.~\ref{fig:three_type} (a). Another emerging approach in the context of large language models utilizes text features and contrastive loss, as shown in Fig.~\ref{fig:three_type} (b-c).
\textbf{(1)} For the \textbf{standard paradigm}, the training objective (\textit{i.e.}, cross-entropy loss with softmax) of one point can be written as:
\begin{equation}
    \mathcal{L}_{ce} = -\log(\frac{\exp(z_+)}{\sum_{j=1}^k \exp \left(z_j\right)}),
    \label{eq:ce_loss}
\end{equation}
where $\boldsymbol{z}  $=$ (z_1, z_2, \dots, z_k) \in \mathbb{R} ^k$ represents the predicted scores (logits) for $k$ classes, and its ground-truth(GT) label is a one-hot vector $\boldsymbol{Z} \in \mathbb{R} ^k$. The objective is to maximize $z^+$ (\textit{i.e.}, the predicted score for the GT class). 
\textbf{(2)} For the \textbf{text contrastive paradigm}, the training objective (\textit{i.e.}, InfoNCE loss) of one point can be formulated as:
\begin{equation}
    \mathcal{L}_{con} = -\log(\frac{\exp(\boldsymbol{p}^{\top} \cdot \boldsymbol{t}_+ / \tau)}{\sum_{j=1}^m \exp \left(\boldsymbol{p}^{\top} \cdot \boldsymbol{t}_j / \tau\right)}),
    \label{eq:con_loss}
\end{equation}
where $\boldsymbol{p}$$\in$$\mathbb{R}^d$ represents the learned feature of a point, $(\boldsymbol{t}_1^{\top}, \boldsymbol{t}_2^{\top}, \dots, \boldsymbol{t}_m^{\top}) \in \mathbb{R} ^{m \times d}$ denotes the features of $m$ texts extracted by the text encoder, and $d$ is the feature dimension of the point cloud and texts. The term $\boldsymbol{p}^{\top} \cdot \boldsymbol{t}$ indicates the similarity score between the point cloud and texts. $\tau$ is a fixed temperature coefficient. The objective is to maximize $\boldsymbol{p}^{\top} \cdot \boldsymbol{t}_+$, which represents the feature similarity between the point cloud and the text of its GT category.
Based on the formulation, we derive the following three observations.

\textbf{ $\bullet$ From the $k$ vs. $m$ perspective:} 
In the standard paradigm, k represents the number of categories, typically set to 20 (\textit{e.g.}, ScanNet20) or 200 (\textit{e.g.}, ScanNet200). It is also the number of output channels of the segmentation head. This means that k is task-specific and fixed, preventing it from simultaneously supporting segmentation with 20 and 200 categories.
Conversely, in the text contrastive paradigm, $m$ represents the number of texts used for calculating similarity, and it only needs to be the same within one batch. Hence, $m$ is flexible and variable. This paradigm allows the same model to train and infer across 20, 200, or more categories.
\textbf{Observation 1:} Text contrastive training enhances the flexibility of point cloud segmentation in two aspects. \textbf{\romannumeral1)} One model supports segmentation with different category-set. \textbf{\romannumeral2)} Theoretically, it supports any text query for inference, not limited to predefined categories.

\textbf{$\bullet$ From the $z$ vs. $\boldsymbol{p}^{\top} \cdot \boldsymbol{t}$ perspective:} 
Ignoring $\tau$, Eq.~\eqref{eq:ce_loss} and Eq.~\eqref{eq:con_loss} become equivalent, where $z$ is the prediction score (real number) of the segmentation head for a specific category, and $\boldsymbol{p}^{\top} \cdot \boldsymbol{t}$ is the similarity (real number) between the point cloud feature and the feature of a specific text.
However, the distinction emerges in supervision: the standard paradigm uses one-hot vectors $\boldsymbol{t} \in \mathbb{R} ^d$ of class labels, while the text contrast paradigm employs class-specific text feature vectors $\boldsymbol{t} \in \mathbb{R} ^d$. The one-hot vectors of different categories are \textbf{discrete and orthogonal}, facilitating independent feature training. Conversely, the text feature vectors of different categories are \textbf{continuous and correlated}, allowing the contrastive training to yield point cloud features with semantic meaning but potentially introducing ambiguity.
\textbf{Observation 2:} 
Text contrastive training does not always guarantee improved point cloud learning. \textbf{\romannumeral1)} The standard paradigm is more efficient when recognizing a small number of categories with high separability (\textit{e.g.}, low category correlation in ScanNet20), thanks to its independent one-hot vectors. \textbf{\romannumeral2)} For recognizing numerous correlated fine-grained categories (\textit{e.g.}, ``office chair," ``armchair," ``sofa chair" subcategories in ScanNet200), the text contrast paradigm, supervised by continuous semantic concepts, excels.

\textbf{$\bullet$ From the $d$ perspective:} 
In the text contrastive paradigm, $d$ is the dimension for text and point cloud features. The feature dimension of point clouds at the final layer of segmentation tasks is typically low. One choice is to project the original 512 or 768-dimensional text features to a lower dimension (\textit{e.g.}, $d$=$64$) for contrastive training with point cloud features. However, this projection compromises the powerful prior of the text encoder to fit the current training set. 
Alternatively, text features from the text encoder remain unchanged, while point cloud features project into a higher dimension (\textit{e.g.}, $d$=$512$). Although it preserves prior text knowledge, dense point cloud segmentation tasks become computationally expensive. Moreover, high-dimensional text features might suffer from redundancy or ambiguity, hindering discriminative point cloud feature learning.
\textbf{Observation 3:} 
\textbf{\romannumeral1)} Projecting text features to lower dimensions risks losing embedded prior knowledge and overfitting the training set, limiting generalization. 
\textbf{\romannumeral2)} Keeping original text dimensions burdens point cloud processing and may yield ambiguous, less distinctive features.

Given the above observations, our goal is to maximize text-contrastive training benefits while minimizing drawbacks. \textbf{\romannumeral1)} To acquire more diverse and discriminative textual features $\boldsymbol{t}$, we propose employing LLM-based text enrichment (Sec.~\ref{sec:text}). \textbf{\romannumeral2)} To achieve dimensionality reduction and de-redundancy without compromising text priors, we introduce statistical-based significant feature selection (Sec.~\ref{sec:selection}).

\begin{figure*}[!htbp]
\centering
\setlength{\abovecaptionskip}{2pt}
\includegraphics[width=1.0 \textwidth]{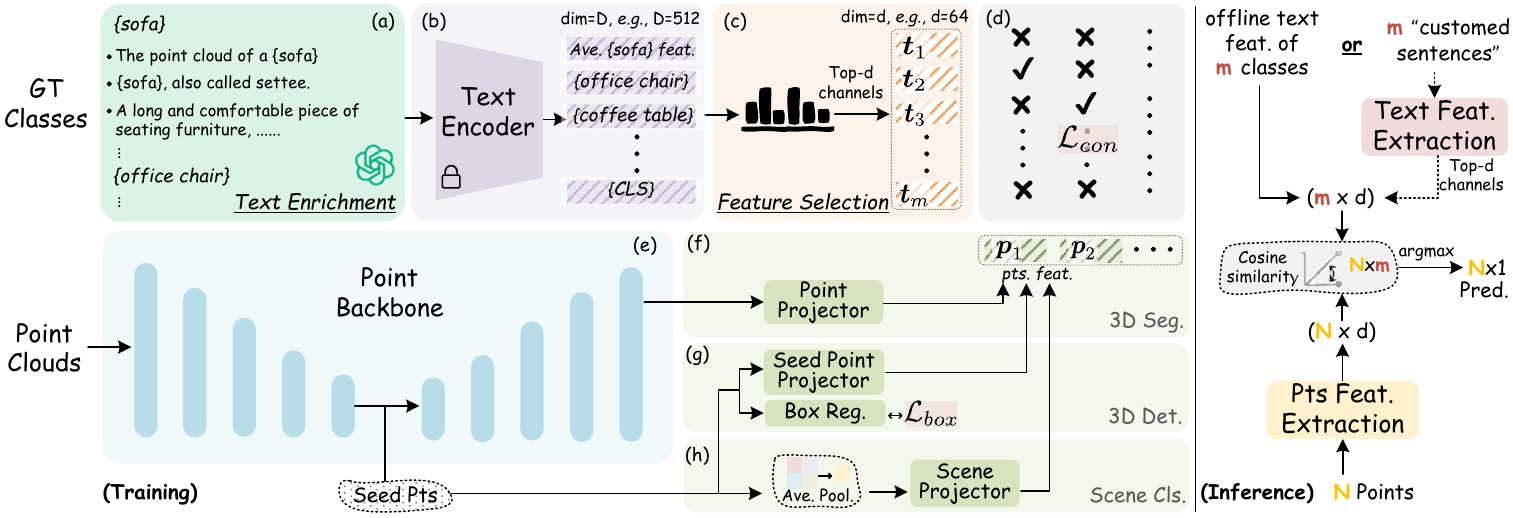} 
\caption{
Model Architecture. Left: Training process. (a) Diversified fine-grained descriptions generated by LLM for GT categories, achieving text enrichment. (b) Extracting features from the generated text using a frozen text encoder and obtaining averaged features for multiple sentences of the same category. (c) Statistical-based significance feature selection for text feature dimensional reduction. (e) Point cloud backbone extracting dense point cloud features. (f) Semantic segmentation. (g) 3D object detection. (h) 3D scene type classification. Projector layers transform point features into the same dimension as text features. (d) Text-point cloud contrastive loss. Right: Inference process, illustrated by the segmentation task.
}
\label{fig:Framework}
%\vspace{-15pt}
\end{figure*}

\subsection{Model Overview}
\label{sec:overview}
We aim to leverage the powerful prior of large pre-trained language models to assist in feature learning for data-scarce point clouds, enabling the acquisition of more efficient and semantically meaningful features. As shown in Fig.~\ref{fig:Framework}, the model can be divided into point cloud feature extraction and text feature extraction.
In the point part, we adopt a U-Net style backbone (Fig.~\ref{fig:Framework}~(e)) to extract dense point cloud features or use an encoder with downsampling structures to extract sampled sparse seed point features, for segmentation (Fig.~\ref{fig:Framework}~(f)), detection (Fig.~\ref{fig:Framework}~(g)), and scene classification (Fig.~\ref{fig:Framework}~(h)).
Unlike standard methods, we discard the one-hot label-based segmentation head. Instead, we employ a Point Projector composed of MLPs to match the dimensional-reduced text features for contrastive training.

In the text part, we first utilize the Large Language Model (LLM) to generate fine-grained text descriptions based on the GT categories of the point cloud to enrich semantic concepts (see Fig.~\ref{fig:Framework}~(a) and Sec.~\ref{sec:text}). Subsequently, the generated text undergoes a frozen text encoder (the text encoder of CLIP is used in our implementation) to extract features (Fig.~\ref{fig:Framework}~(b)).
Given the ambiguity and redundancy in high-dimensional text features, we use a statistical approach for significant feature selection, reducing dimensionality (see Fig.~\ref{fig:Framework}~(c) and Sec.~\ref{sec:selection}). This training-free module avoids the computation load of projecting point cloud features into high dimensions and maintains text priors by preventing text features from being projected into low dimensions.

\subsection{LLM-Based Text Enrichment}
\label{sec:text}
The categories of the point cloud can naturally be converted into text and fed into the text encoder to obtain text embeddings for supervision. However, due to the continuity of the text embedding space, there could be similarities and correlations between different category embeddings. To obtain more discriminative text features and to distinguish finer-grained concepts (\textit{e.g., ``sofa chair'' vs. ``armchair'', ``kitchen cabinet'' vs. ``bathroom cabinet''}), we leverage the text generation capability of LLMs to obtain more detailed and comprehensive category descriptions, enriching associated semantic concepts.

We constructed text descriptions using the following methods: \textbf{(1)} Generating sentences with fixed templates like \textit{``The point cloud of \{CLS\}.''}. \textbf{(2)} Querying GPT-3~\cite{brown2020language} for synonyms and constructing sentences like \textit{``\{CLS\}, also sometimes called \{Synonym\}.''}. \textbf{(3)} Utilizing Q\&A-style instructions to ask GPT-3 about the appearance, characteristics, functions, and other attributes of each category to generate detailed answers. In the end, we obtained approximately 15 detailed text descriptions for each category, aiming to thoroughly explore the semantic priors within the text feature space.
Furthermore, we found that finer-grained subcategory concepts are beneficial for feature learning. For instance, in ScanNet20, the category \textit{``chair"} is divided into subcategories such as \textit{``office chair," ``armchair," ``sofa chair,"} in ScanNet200. Therefore, we propose a variant for ScanNet20: joint training of 20 and 200 categories to learn more detailed semantic representations.

\subsection{Significant Feature Selection}
\label{sec:selection}
High-dimensional textual features add burdens to point cloud projections and may lead to redundancy and ambiguity. Additionally, trainable projection layers for dimension reduction compromise the inherent textual priors and fit the training set. Inspired by the concept of APE~\cite{zhu2023not} for refining image features, we propose utilizing training-free statistical-based significance feature selection, achieving dimensionality reduction while preserving textual priors.

We constructed $s$ textual descriptions for each of the $m$ categories, \textit{e.g.}, $\boldsymbol{t}_m^s$ represents the feature of the $s$-th text for the $m$-th category, and $\boldsymbol{t}_m$ denotes the average feature of all texts for the $m$-th category. Our objective is to transform $\boldsymbol{t}_m \in \mathbb{R}^D, e.g., D=512$ into $\boldsymbol{t}_m' \in \mathbb{R}^d, e.g., d=64$ by channel significance ranking. To alleviate feature redundancy among descriptions of the same category, the selected $d$ channels should exhibit \textbf{minimal intra-class similarity}. To differentiate among the $m$ categories, the selected $d$ channels should \textbf{maximize inter-class variance}.

The intra-class similarity ranking can be computed by:
\begin{equation}
    \boldsymbol{S} = \frac{1}{m}\sum_{n=1}^{m}\left ( \frac{1}{s^2} \sum_{i=1}^{s} \sum_{j=1}^{s} \boldsymbol{t}_n^i \odot \boldsymbol{t}_n^j \right ) ,
    \label{eq:similarity}
\end{equation}
where $\boldsymbol{S} \in \mathbb{R}^D$ represents the significance ranking of the $D$ channels in terms of intra-class similarity. 
The inter-class variance can be calculated using channel-wise variance:
\begin{equation}
    \boldsymbol{V}=\sigma(\mathbf{T}), \mathbf{T}=\{\boldsymbol{t}_1^{\top}, \boldsymbol{t}_2^{\top}, ..., \boldsymbol{t}_m^{\top}\} \in \mathbb{R} ^{m\times D},
    \label{eq:variance}
\end{equation}
where $\sigma(\cdot)$ is the variance function, and $\boldsymbol{V} \in \mathbb{R}^D$ represents the significance ranking of the $D$ channels in terms of inter-class variance.
Finally, we obtain the overall significance ranking of the $D$ channels as follows:
\begin{equation}
    \boldsymbol{R}=\lambda \boldsymbol{V} - (1-\lambda )\boldsymbol{S}.
    \label{eq:ranking}
\end{equation}
By selecting the top-$d$ channels, we obtain the reduced text features $\boldsymbol{t}'$$\in$$\mathbb{R}^d$ for training. $\lambda$$=$$0.7$ in our implementation.

\subsection{Training and Inference}
\label{sec:training_inference}
In training, the text part, including text generation, feature extraction, and low-dimensional feature selection, can all be performed offline. Hence, the primary computational burden lies in the point part, which is comparable to standard networks. In segmentation and 3D scene classification tasks, contrastive loss $\mathcal{L}_{con}$ (Eq.~\eqref{eq:con_loss}) is employed for supervision. For the detection task, apart from the contrastive loss, additional losses related to box regression $\mathcal{L}_{box}$ are also utilized.

The inference process is shown in the right part of Fig.~\ref{fig:Framework}. Offline-acquired text features are directly compared with the point cloud features, computing the cosine similarity, and the score with the highest is considered the prediction result. Additionally, the text-encoder-required mode is supported to accepts user-customed text, including synonyms or detailed descriptions for categories, enabling more natural human-computer interaction.

%%%%%%%%%%%%%%%%%%%%%%%%%%%%%%%%%%%%%%%%%%%%%%%%%%%%%%%
% 4. Experiments
%%%%%%%%%%%%%%%%%%%%%%%%%%%%%%%%%%%%%%%%%%%%%%%%%%%%%%%
\section{Experiments}
\label{sec:exp}

\begin{table*}[htbp]
\begin{minipage}[t]{0.6\textwidth}
    \footnotesize
    \centering
    \setlength{\abovecaptionskip}{2pt}
    \setlength{\tabcolsep}{1.5pt}
    \renewcommand{\arraystretch}{1.0}
    \begin{tabular}{c|cccc}
    \toprule
    Method       & \scriptsize{ScanNet20} & \scriptsize{ScanNet200\;\;} & \scriptsize{ScanNet485} & \scriptsize{S3DIS13}    \\  \midrule
    Mink.Net~\scriptsize{\cite{choy20194d}}               & 72.2                  & 25.1                  & -            & 65.4          \\
    Mix3D~\cite{nekrasov2021mix3d}                  & 73.6                & -                & -            & -             \\
    LargeKernel3D~\cite{chen2023largekernel3d}          & 73.2            & -                & -            & -             \\
    \scriptsize{PointConvFormer~\cite{wu2023pointconvformer}}        & 74.5            & -                & -            & -             \\
    PointTransformer~\cite{zhao2021point}       & 70.6                    & -                & -            & 70.4          \\
    Stratified Trans.~\cite{lai2022stratified} & 74.3                & -                & -            & \textbf{72.0}            \\
    FastPointTrans.~\cite{park2022fast}   & 72.4                     & -                & -            & 70.4          \\
    LG~\scriptsize{\cite{rozenberszki2022language}}            & -            & 28.9          & -            & -             \\
    CeCo~\cite{zhong2023understanding}                   & -              & 32.0            & -            & -             \\
    OctFormer~\cite{Wang-2023-octformer}              & 74.5              & 31.7          & -            & -             \\
    SWin3D-L~\cite{yang2023swin3d}               & 74.2                   & -                & -            & 69.6          \\
    PointTransformerV2~\scriptsize{\cite{wu2022point}}     & 74.8 (75.4)               & 31.9             & 13.3            & 71.1 (71.6)          \\ \midrule
    LAST-PCL (Ours)                   & 73.3              & \textbf{33.3}            & \textbf{14.3}        & 70.8 (71.3) \\
    LAST-PCL (Ours)\dag \scriptsize{~(joint 20-200)}                  & 75.1 (\textbf{75.9})            & 29.5                 & -            & -             \\ \bottomrule
    \end{tabular}
    \caption{
    Performance comparison on four semantic segmentation benchmarks: ScanNet20, ScanNet200, ScanNet485, and S3DIS13 (20, 200, 485, and 13 categories, respectively). Voted results are shown in parentheses (), with the method consistent with PointTransformerV2. \dag~ indicates joint training on ScanNet20 and ScanNet200 with evaluation across both benchmarks.
    }
    \label{tab:seg}
  \label{tab:tableA}
  %\vspace{-15pt}
\end{minipage}%
\begin{minipage}{0.01\textwidth}
${ }$
\end{minipage}%
\begin{minipage}[t]{.4\textwidth}
    \footnotesize
    \centering
    \setlength{\abovecaptionskip}{2pt}
    \setlength{\tabcolsep}{1.7pt}
    \begin{tabular}{c|cc|cc}
    \toprule
    \textbf{Method} & \textbf{\scriptsize{Pre.}} & \textbf{\scriptsize{Mul.}} & \textbf{\scriptsize{@0.25}} & \textbf{\scriptsize{@0.5}} \\ 
    \midrule
    VoteNet~\cite{qi2019deep}         & \ding{55}              & \ding{55}              & 62.9              & 39.9             \\
    % MLCVNet~\cite{xie2020mlcvnet}         & \ding{55}              & \ding{55}              & 64.5              & 41.4             \\
    H3DNet~\cite{zhang2020h3dnet}          & \ding{55}              & \ding{55}              & 64.4              & 43.4             \\
    \scriptsize{3DETR~\cite{misra2021end}}           & \ding{55}              & \ding{55}              & 65.0                & 47.0               \\
    GroupFree~\cite{liu2021group}       & \ding{55}              & \ding{55}              & 66.3              & 48.5             \\
    % PointBERT~\cite{yu2022point}       & \ding{51}             & \ding{55}              & 61.0                & 40.2             \\
    Point-MAE~\cite{pang2022masked}        & \ding{51}             & \ding{55}              & 62.1              & 41.2             \\
    Point-M2AE~\cite{zhang2022point}        & \ding{51}             & \ding{55}              & 66.3              & 48.3             \\
    ACT~\cite{dong2023act}             & \ding{51}             &  \ding{55}             & 63.8              & 42.1             \\
    ULIP~\cite{xue2023ulip}            & \ding{51}             & T, I   & -                 & 49.6             \\
    Text4Point~\cite{huang2023joint}      & \ding{55}              & T          & -                 & 39.7             \\
    Distillation~\cite{yao20223d}    & \ding{55}              & I         & 63.4              & 42.2             \\
    I2P-MAE~\cite{zhang2023learning}        & \ding{51}             & I              & 63.9              & 43.1             \\
    BUTD-DETR~\cite{jain2022bottom}       & \ding{55}              & T          & 63.0                & 43.8             \\
    EDA~\cite{wu2023eda}             & \ding{55}             & T          & 64.1              & 45.3             \\
    \midrule
    LAST-PCL (Ours)            & \ding{55}              & T          & \textbf{67.3}              & \textbf{50.1}             \\ \bottomrule
    \end{tabular}
    \caption{Comparison of 3D object detection performance on ScanNet. `Pre.' indicates self-supervised pretraining. `Mul.' signifies contrastive supervision using multimodal images (I) or text (T). Our method reports the average of 6 training results.}
    \label{tab:det}
    %\vspace{-15pt}
\end{minipage}
\end{table*}

\begin{table}[t]
\small
\centering
\setlength{\abovecaptionskip}{2pt}
\setlength{\tabcolsep}{8pt}
\resizebox{\columnwidth}{!}{
\begin{tabular}{c|cc}
\toprule
\textbf{Method} & \textbf{Acc.} & \textbf{mIoU} \\ 
\midrule
Pointnet++~\cite{qi2017pointnet++}      & 84.8          & 69.1          \\
DGCNN~\cite{wang2019dynamic}           & 77.0            & 62.1          \\
MinkowskiNet\scriptsize{~\cite{choy20194d}}         & 87.4           & 70.6      \\
Multi-task~\scriptsize{~\cite{huang2020indoor}}      & 90.3          & 77.2          \\
PointTransformerV2~\cite{wu2022point}            & 89.8          & 79.4          \\ \midrule
LAST-PCL (Ours)            & \textbf{91.0}            & \textbf{81.3}          \\ 
\bottomrule
\end{tabular}
}
\caption{Comparison of 3D scene classification performance on the ScanNet dataset.}
\label{tab:cls}
%\vspace{-15pt}
\end{table}

%%%%%%%%%%%%%%%% 4.1 Segmentation %%%%%%%%%%%%%%%
\subsection{3D Semantic Segmentation}
\label{sec:exp_seg}
\textbf{Settings:} 
We evaluate our approach using two datasets, ScanNet~\cite{dai2017scannet} (1613 indoor room scans, with 312 for validation) and S3DIS~\cite{armeni20163d} (272 rooms from 6 areas, with Area-5 as validation), across four semantic segmentation benchmarks: (1) \textbf{ScanNet20} with 20 common semantic labels; (2) \textbf{ScanNet200}~\cite{rozenberszki2022language} with 200 rich semantic labels, including finer-grained classifications and novel categories beyond the initial 20; (3) 
\textbf{ScanNet485}, a more challenging benchmark with 485 semantic labels introduced by BUTD-DETR~\cite{jain2022bottom} for detection, is employed for the first time for segmentation in our work, and we retrained PointTransformerV2 on this benchmark accordingly;
and (4) \textbf{S3DIS13}, featuring 13 semantic labels. Mean IoU served as the evaluation metric. Our model utilized PointTransformerV2 as the point backbone, employing the same hyperparameters, data augmentation, and training epochs.

\textbf{Results:}
\textbf{\romannumeral1)} 
As shown in the second-last row of Tab.~\ref{tab:seg}, overall, the proposal LAST-PCL achieves SOTA or comparable performance, demonstrating the potential of leveraging text priors of LLMs to facilitate point cloud feature learning.
\textbf{\romannumeral2)} LAST-PCL exhibits performance degradation compared to the baseline on the ScanNet20 and S3DIS13 benchmarks. However, on the more challenging ScanNet200 and ScanNet485 benchmarks, our method shows significant performance improvements, validating our Observation 2 in Sec.~\ref{sec:Theoretical_Analysis}. In cases with a small number of categories, there is significant diversity between classes, favouring efficient supervision with discrete, and orthogonal one-hot labels. Conversely, for larger category sets with highly interrelated and similar, continuous and semantically meaningful text features provide a superior prior.
\textbf{\romannumeral3)} Notably, LG, also based on text-guided contrastive training, shows inferior performance. This deficiency can be attributed to the lack of exploration into finer-grained textual descriptions and the retention of the original 512-dimensional text features, leading to redundancy and ambiguity.
\textbf{\romannumeral4)} 
Moreover, we trained a variant to illustrate the flexibility of our approach. As shown in the last row of Tab.~\ref{tab:seg}, joint training on ScanNet20 and ScanNet200 led to SOTA performance on ScanNet20. The result demonstrates the advantage of learning more fine-grained semantic concepts for point cloud understanding. Unlike conventional methods requiring two segmentation heads with varying channel numbers for training on two benchmarks, we employ a unified network for both tasks.

%%%%%%%%%%%%%%%% 4.2 Detection %%%%%%%%%%%%%%%
\subsection{3D Object Detection}
\label{sec:exp_det}
\textbf{Settings:} 
We utilize the ScanNet detection benchmark to evaluate the detection performance of our method, with mAP@0.25 and mAP@0.5 as metrics. We employ the encoder of EDA as the backbone for the detection task, with consistent hyperparameters and augmentation as in EDA.

\textbf{Results:}
\textbf{\romannumeral1)} As shown in Tab.~\ref{tab:det}, LAST-PCL achieves SOTA performance and offers significant improvements compared to the baseline. This result validates the effectiveness of our text-contrastive training method for the detection task.
\textbf{\romannumeral2)} 
With the recent advancements in large multimodal models, some studies attempt to distill prior features from pre-trained text and image encoders and transfer them to point cloud tasks (indicated as T for text and I for image in Tab.~\ref{tab:det}). Our method outperforms these multimodal training approaches thanks to text enrichment and feature selection.

%%%%%%%%%%%%%%%% 4.3 Classification %%%%%%%%%%%%%%%
\subsection{3D Scene Classification}
\label{sec:exp_cls}
\textbf{Settings:} 
Identifying 3D room types is crucial for intelligent robot tasks, serving as a measure of 3D understanding. The ScanNet dataset we employed comprises 21 room types (\textit{e.g., ``Kitchen", ``Bedroom", ``Lounge"}). We adopt the same dataset split and metrics (Acc. and mIoU) as Multi-task~\cite{huang2020indoor}. The encoder of PointTransformerV2 serves as the point backbone. Notably, tens of thousands of points from one scene are ultimately downsampled and pooled into a 512-dimensional feature vector. Further dimensionality reduction could lead to information loss. A single 512-dim vector per scene does not significantly increase computational complexity. Therefore, in this task, we refrain from reducing the dimensionality of text features. Nevertheless, we applied the proposed feature selection method to rerank feature channels.

\textbf{Results:}
\textbf{\romannumeral1)} As depicted in Tab~\ref{tab:cls}, our method achieves SOTA performance, surpassing Multi-task, which necessitates more supervision and computational resources through joint semantic segmentation and scene classification training. 
\textbf{\romannumeral2)} Remarkably, our approach even outperforms PointTransformerV2 supervised with one-hot labels, which appears to contradict findings from segmentation tasks concerning small-scale categories. We attribute this to the high semantic similarity existing among room types. For instance, different room types exhibit object, visual, and functional similarities. Supervising with continuous, semantically meaningful text features proves to be more effective.

%%%%%%%%%%%%%%%% 4.4 Ablation %%%%%%%%%%%%%%%
\begin{table}[t]
\small
\centering
\setlength{\abovecaptionskip}{2pt}
\setlength{\tabcolsep}{2pt}
\resizebox{\columnwidth}{!}{
\begin{tabular}{lcc|cccc}
\toprule
    & Temp. & Detail & ScanNet20 & ScanNet200 & S3DIS & SceneCls \\ \midrule
\#1 & \ding{51}     &      & 75.0      & 31.5      & 70.1 & 80.9    \\
\#2 & \ding{51}     & \ding{51}      & \textbf{75.1} \scriptsize{(0.1 $\uparrow$)}     & \textbf{33.3} \scriptsize{(1.8$\uparrow$)}      & \textbf{70.8} \scriptsize{(0.7$\uparrow$)} & \textbf{81.3} \scriptsize{(0.4$\uparrow$)}    \\ \bottomrule
\end{tabular}
}
\caption{Ablation study of Text Enrichment across different datasets and tasks. ``Temp.'' denotes sentence construction using templates. ``Detail'' signifies fine-grained descriptions generated by GPT. Measured by mIoU.}
\label{tab:ab_text}
%\vspace{-15pt}
\end{table}

\begin{table}[t]
\small
\centering
\setlength{\abovecaptionskip}{2pt}
\setlength{\tabcolsep}{5pt}
\resizebox{\columnwidth}{!}{
\begin{tabular}{cc|cc|c}
\toprule
 &  \multicolumn{1}{l|}{Method} & ScanNet20 & ScanNet200    & Characteristic \\
\midrule
\#1                   & 512 (Orig.)                & 74.8   & 30.4       & \faHourglassStart ~\faLowVision ~\faCopy      \\
\#2                   & 64 (Proje.)                  & 75.0   & 31.8        & \faUnlink       \\
\#3                   & 64 (Select)                 & \textbf{75.1}   & \textbf{33.3}     & \faThumbsUp \\
\bottomrule
\end{tabular}
}
\caption{Performance comparison of different text feature transformation methods. ``Orig.'': preserving the original dimensions. ``Proje.'': dimension transformation using learnable projection layers. ``Select'': the channel-wise feature selection method we employed. Measured by mIoU.}
\label{tab:feat_select}
%\vspace{-15pt}
\end{table}

\subsection{Ablation Studies}
\label{sec:exp_ab}
\textbf{Text Enrichment.}
To validate the role of fine-grained textual descriptions, we present the ablation experiments as shown in Tab~\ref{tab:ab_text}. 
\textbf{\romannumeral1)} In case \#1, we only use fixed templates to generate descriptions, such as \textit{``The point cloud of a \{CLS\}"}. Remarkably, this already yields substantial performance, indicating that the language model provides powerful and discriminative semantic priors. 
\textbf{\romannumeral2)} With the addition of elaborative descriptions generated by GTP, performance further improves. We showcase results across different datasets and tasks to demonstrate that this enhancement is not by chance. 
\textbf{\romannumeral3)} On the more challenging ScanNet200 segmentation benchmark, providing more detailed semantic descriptions leads to significant performance gains. This is also intuitively aligned, as ScanNet200 comprises numerous semantically similar categories and finer-grained descriptions aid in differentiation. In contrast, the other benchmarks (with 20, 13, and 21 categories, respectively) receive sufficient discriminative features from text encoder templates, lessening the impact of additional descriptions.

\textbf{Feature Selection.}
To highlight the distinction between our text feature utilization and standard methods, in Tab.~\ref{tab:feat_select}, our method (case \#3), while achieving comprehensive performance leadership (\faThumbsUp), also offers the following advantages: 
\textbf{\romannumeral1)} Computational efficiency: In case \#1, preserving original high-dimensional features extracted from the text encoder requires projecting point clouds to high dimensions to align with the text, resulting in significant computational expenses (\faHourglassStart) for dense point cloud prediction.
\textbf{\romannumeral2)} Avoidance of feature ambiguity (\faLowVision): Fine-grained descriptions contain rich semantics, particularly for ScanNet200, with significant semantic similarity among categories. High-dimensional raw features struggle to differentiate closely related categories. Our method demonstrates a 2.9\% performance improvement over case \#1 on ScanNet200, indicating the benefit of channel-wise feature selection for isolating more distinctive features. 
\textbf{\romannumeral3)} Reducing feature redundancy: On the ScanNet20, despite these benchmarks having only 20 categories and low inter-feature similarity, our use of just 1/8 of the feature channels attains higher performance than case \#1. This suggests redundancy (\faCopy) among high-dimensional features. 
\textbf{\romannumeral4)} Preservation of original textual priors: Case \#2 trains a 512-to-64 projection layer to achieves dimensionality reduction. However, this projection layer migrates the original textual priors to the feature space fitted to the training set, compromising the pre-trained textual priors (\faUnlink) and limiting model generalization.

%%%%%%%%%%%%%%%%%%%%%%%%%%%%%%%%%%%%%%%%%%%%%%%%%%%%%%%
% 6. Discussion and Analysis
%%%%%%%%%%%%%%%%%%%%%%%%%%%%%%%%%%%%%%%%%%%%%%%%%%%%%%%
\section{Discussion and Analysis}
\label{sec:analysis}
In this section, we conduct experiments to validate the observations presented in Sec.~\ref{sec:Theoretical_Analysis}.

\begin{figure}[t]
    \centering
    \setlength{\abovecaptionskip}{2pt}
    \begin{minipage}{0.25\textwidth}
        \centering
        \includegraphics[width=\textwidth]{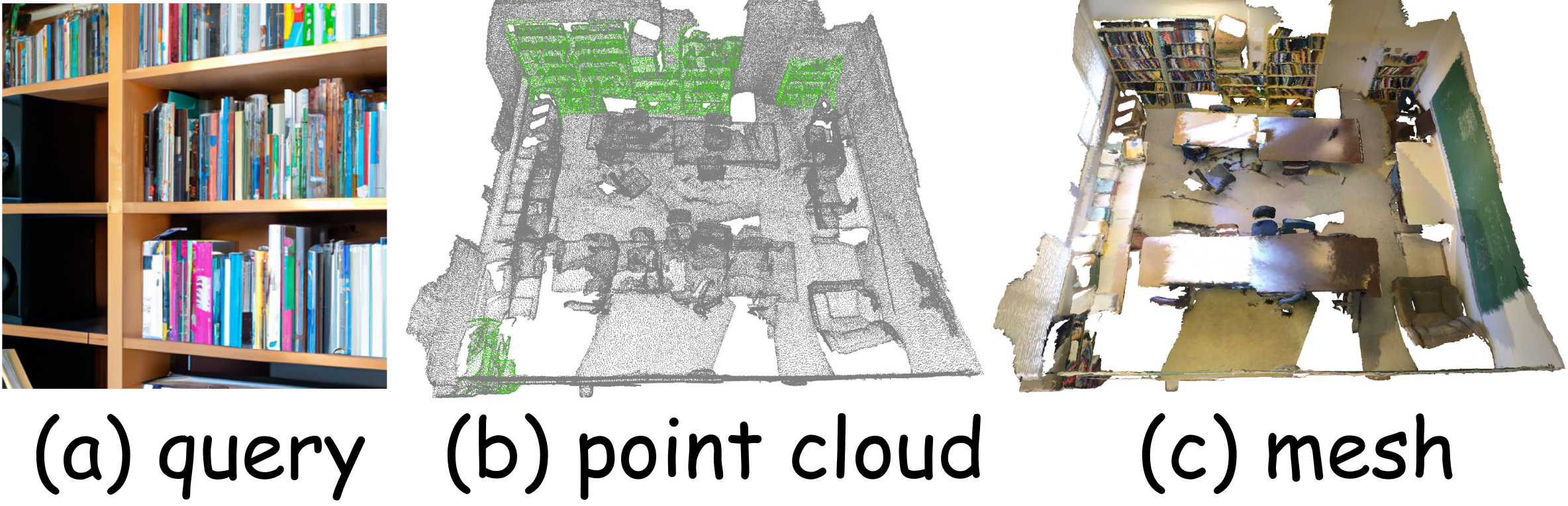} 
        \caption{(a) Query images generated by DALL·E~\cite{ramesh2021zero}. (b) Coloured regions show highly similar points to the query image. (c) Reference 3D mesh for readers.}
        \label{fig:image_query}
    \end{minipage}
    { } % gap
    \begin{minipage}{0.20\textwidth}
        \centering
        \setlength{\abovecaptionskip}{2pt}
        \small
        \setlength{\tabcolsep}{1pt}
        \resizebox{\columnwidth}{!}{
            \begin{tabular}{c|ccc}
                \toprule
                Method      & mIoU  & mAcc  & allAcc    \\
                \midrule
                64 (\scriptsize{Proje.}) & 4.0  & 14.4 & 29.1 \\
                64 (\scriptsize{Select}) & \textbf{12.5} & \textbf{17.7} & \textbf{45.0} \\
                \bottomrule
            \end{tabular}
        }
        \caption{Generalization comparison on S3DIS, with models trained on ScanNet20. Project-based method shows significant decline.}
        \label{tab:gener}
    \end{minipage}
\end{figure}

\textbf{Analysis 1: Flexibility of Text Contrastive Training.}
\textbf{1)} Due to the absence of constraints imposed by category-related segmentation heads, our model accommodates joint training on datasets with arbitrary sets of categories as long as the dimension of the point cloud features matches that of the text features. Detailed in the supplementary material (Tab.~\ref{tab:joint_traing}), we joint-train on multiple datasets, enabling a unified model for diverse benchmark inference.
\textbf{2)} In the supplementary material (Tab.~\ref{tab:text_query}), we demonstrate that using different LLMs to randomly generate text descriptions for categories as queries yields comparable performance, indicating that our model learned meaningful semantic concepts.
\textbf{3)} 
Furthermore, we employ the text encoder from CLIP, whose feature space is already aligned with image features. Our model supervises the alignment of point cloud features with text features, indirectly achieving alignment with image features, even though we did not use any images for training. Therefore, LAST-PCL also possesses the potential to use images as queries. As illustrated in Fig.~\ref{fig:image_query}, input query images undergo feature extraction through CLIP's image encoder, with the coloured regions representing point clouds highly similar to the image. This also indirectly validates that our feature selection method achieves dimensionality reduction without compromising the underlying text priors.

\begin{figure}[t]
\centering
\setlength{\abovecaptionskip}{2pt}
\includegraphics[width=0.47 \textwidth]{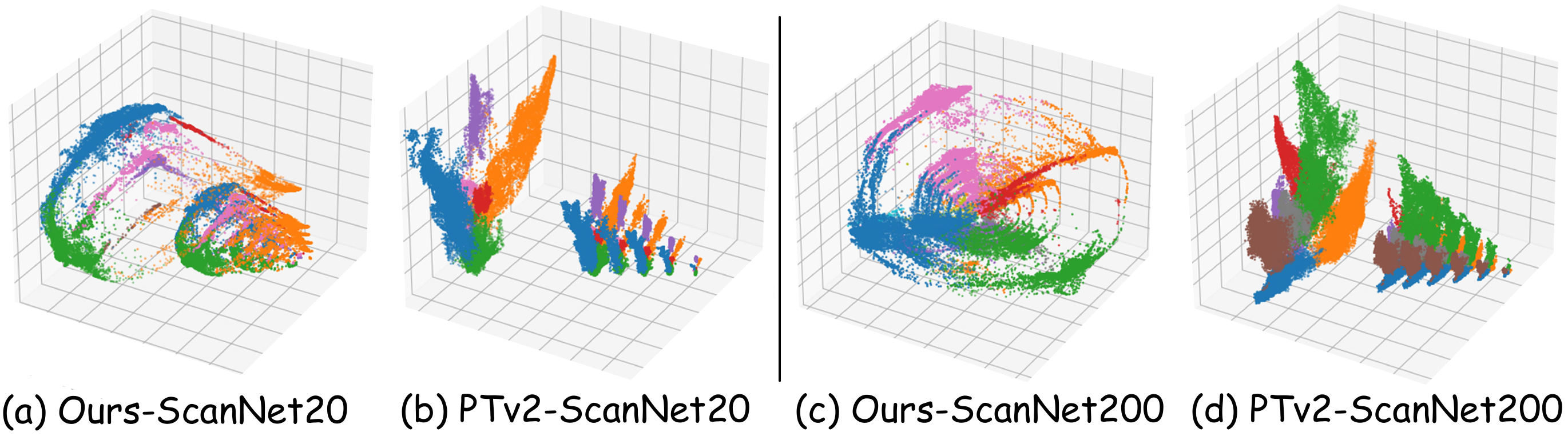} 
\caption{
PCA visualization of features. Colors depict categories. The features learned by ours exhibit continuity, unlike the discreteness of PointTransformerV2.}
\label{fig:feat_vis}
\vspace{-10pt}
\end{figure}

\textbf{Analysis 2: Text contrastive training does not always lead to improved accuracy in point cloud understanding tasks.} Despite claiming that text contrastive enhances the point cloud learning of semantically meaningful features, our approach exhibits a slight performance decline compared to the baseline on some small-scale category-set benchmarks, such as ScanNet20 and S3DIS13, as shown in Tab.~\ref{tab:seg}. However, our method demonstrates significant superiority over the baseline on benchmarks with larger category sets. To explore the reasons, we conducted PCA~\cite{minka2000automatic} visualization on the point cloud features on ScanNet20 and ScanNet200, as depicted in Fig.~\ref{fig:feat_vis}. Notably, the features of different categories in PTv2 are discrete and orthogonal (Fig.~\ref{fig:feat_vis}(b, d)), aligning with the characteristics of one-hot labels.
In contrast, the features learned by our method exhibit continuity (Fig.~\ref{fig:feat_vis}(a, c)), consistent with the semantic correlations among distinct text features. For tasks involving small-scale category sets, the discrete and orthogonal nature of one-hot labels proves more efficient due to the low inter-category correlation. Conversely, supervision of continuous text features is more effective for tasks involving larger category sets, as it provides prior knowledge about the semantic correlations among categories.

\textbf{Analysis 3: The choice of text feature dimensions influences the accuracy and generalization ability of the model.} As analyzed in Tab.~\ref{tab:ab_text}, the original high-dimensional text features not only increase point cloud computation costs but also introduce redundancy and ambiguity. On the other hand, transforming them to lower dimensions through projection layers would disrupt the text priors to fit the training set. To demonstrate this point, we trained models using two dimensionality reduction methods on ScanNet20 and tested them on S3DIS. As shown in Tab.~\ref{tab:gener}, the performance loss from dimensionality reduction through the projection layer is significant, indicating that the projection layer overfits the training set, resulting in reduced generalization capability.

%%%%%%%%%%%%%%%%%%%%%%%%%%%%%%%%%%%%%%%%%%%%%%%%%%%%%%%
% 5. Conclusion
%%%%%%%%%%%%%%%%%%%%%%%%%%%%%%%%%%%%%%%%%%%%%%%%%%%%%%%
\section{Conclusion}
In this paper, we propose LAST-PCL, a method for language-assisted point cloud feature learning. By leveraging text enrichment based on LLM, we enhance the representation of category-relevant semantic concepts. Through statistical-based significant feature selection, we achieve de-redundancy and feature dimensionality reduction without compromising text priors. Extensive experiments on 3D semantic segmentation, object detection, and scene classification tasks demonstrate that LAST-PCL achieves SOTA or comparable performance while learning semantically meaningful point cloud features. 
Importantly, we anticipate that our observations (Sec.~\ref{sec:Theoretical_Analysis}) and validations (Sec.~\ref{sec:analysis}) regarding the impact of text-contrastive training on point cloud learning can provide insightful considerations to the field.

%%%%%%%%% supp %%%%%%%%
\include{supp}

%%%%%%%%% ref %%%%%%%%
\bibliography{aaai24}

\end{document}

%% file: supp.tex
%%%%%%%%%%%%%%%%%%%%%%%%%%%%%%%%%%%%%%%%%%%%%%%%%%%%%%

\clearpage
\newcounter{counter}[section]
\twocolumn[{%
\renewcommand\twocolumn[1][]{#1}%
\begin{center}
    \Large
    \textbf{Supplementary Material for \\ Language-Assisted 3D Scene Understanding}
    \\[20pt]

    % author info
    \large
    Yanmin Wu\textsuperscript{\rm 1} \quad ~
    Qiankun Gao\textsuperscript{\rm 1} \quad ~
    Renrui Zhang\textsuperscript{\rm 2} \quad ~
    Jian Zhang\textsuperscript{\rm 1}$^*$
    \vspace{5pt}
    \\
    %Afiliations
    \textsuperscript{\rm 1} Shenzhen Graduate School, Peking University, China \quad
    \textsuperscript{\rm 2} The Chinese University of Hong Kong, China \\
    % email
    {\tt\small \{wuyanmin, gqk\}@stu.pku.edu.cn} \quad 
    {\tt\small zhangrenrui@link.cuhk.edu.hk} \quad 
    {\tt\small zhangjian.sz@pku.edu.cn}
    \\~
    \\~
\end{center}
}]

In the supplementary material, method details are supplemented in Sec.~\ref{sec:supp_mentod}, including training details and text enrichment methods. In Sec.~\ref{sec:supp_exp}, we add some additional experiments to illustrate the effectiveness of LAST-PCL. Sec.~\ref{sec:supp_vis} provides visualizations of qualitative results.

%%%%%%%%%%%%%%%%%%%%%%%%%%%%%%%%%%%%%%%%%%%%%%%%%%%
% 1. Method Details
%%%%%%%%%%%%%%%%%%%%%%%%%%%%%%%%%%%%%%%%%%%%%%%%%%%
\section{Method Details}
\label{sec:supp_mentod}
% \subsection{Model Details}

\subsection{Derivation of Intra-Class Similarity}
The cosine similarity between two text features $\boldsymbol{t}^i \in \mathbb{R}^D, ~ \boldsymbol{t}^j \in \mathbb{R}^D$ can be defined as:
\begin{equation}
    S_{cos} = \frac{\boldsymbol{t}^i \cdot \boldsymbol{t}^j}{\left | \boldsymbol{t}^i \right | \times \left | \boldsymbol{t}^j \right | } .
\end{equation}
We assume that the text features $\boldsymbol{t}^i, \boldsymbol{t}^j$ have been L2 normalized, and then the cosine similarity can be formulated as:
\begin{eqnarray}
    S_{cos} &=& \boldsymbol{t}^i \cdot \boldsymbol{t}^j   \nonumber\\
      &=& \sum_{c=1}^{D} t^i_c\cdot t^j_c   \nonumber \\
      &=& \sum_{c=1}^{D} (\boldsymbol{t}^i \odot  \boldsymbol{t}^j)_c,
\end{eqnarray}
where $t^i_c$ represents the value of the $c^{th}$ channel of the feature vector $\boldsymbol{t}^i$, and $t^j_c$ similarly. $(\boldsymbol{t}^i \odot  \boldsymbol{t}^j) \in \mathbb{R}^D$ represents the element-wise product of the two feature vectors, indicating the contribution of all $D$ channels to the similarity $S_{cos}$, which can be called the channel significant score vector.

Therefore, the average channel significant score vector of $s$ text features for one category can be expressed as:
\begin{equation}
    \boldsymbol{S} = \frac{1}{s^2} \sum_{i=1}^{s} \sum_{j=1}^{s} \boldsymbol{t}^i \odot \boldsymbol{t}^j.
\end{equation}
The average channel significant score vector of $s$ text features for each of the $m$ categories can be expressed as:
\begin{equation}
    \boldsymbol{S} = \frac{1}{m} \sum_{n=1}^{m}\left ( \frac{1}{s^2} \sum_{i=1}^{s} \sum_{j=1}^{s} \boldsymbol{t}_n^i \odot \boldsymbol{t}_n^j \right ),
\end{equation}
where the vector $\boldsymbol{S}\in \mathbb{R}^D$ stores the contributions of all $D$ channels to the intra-class similarity.

%%%%%%%%%%%%%%%%%%%%%%%%%
\subsection{Our Feature Selection Compared to APE}
\label{sec:com_ape}

Our feature selection method is inspired by APE~\cite{zhu2023not}, which performs dimensionality reduction on image features. We do not only apply it from the image domain to the text domain but also make improvements, as shown in Tab.~\ref{tab:ape_diff}.

% table: comparison with APE
\begin{table}[htbp]
\renewcommand{\arraystretch}{1.0}
\resizebox{\columnwidth}{!}{
\begin{tabular}{cc|c}
\toprule
\multicolumn{1}{c|}{\multirow{5}{*}{$\boldsymbol{S}$}} & APE  & \makecell[c]{Inter-Class Similarity: \\ $\boldsymbol{S}_{inter}$ = $\frac{1}{m^2} \sum\limits_{n_1=1}^{m} \sum\limits_{\substack{n_2=1 \\ n_2 \neq n_1}}^{m} \!\!\! \left ( \frac{1}{s^2} \sum\limits_{i=1}^{s} \sum\limits_{j=1}^{s} \boldsymbol{t}_{n_1}^i \odot \boldsymbol{t}_{n_2}^j \right )$} \\ \cmidrule{2-3} 
\multicolumn{1}{c|}{}                                   & Ours & \makecell[c]{Intra-Class Similarity: \\ $\boldsymbol{S}_{intra} = \frac{1}{m} \sum\limits_{n=1}^{m}\left ( \frac{1}{s^2} \sum\limits_{i=1}^{s} \sum\limits_{j=1}^{s} \boldsymbol{t}_n^i \odot \boldsymbol{t}_n^j \right )$} \\ \midrule 
\multicolumn{2}{c|}{$\boldsymbol{V}$ \small{(APE/Ours)}}               & \makecell[c]{Inter-Class Variance: \\ [2pt] $\boldsymbol{V}\!=\!\sigma(\mathbf{T}), ~ \mathbf{T}\!=\!\left\{\boldsymbol{t}_1^{\top}, \boldsymbol{t}_2^{\top}, \ldots, \boldsymbol{t}_m^{\top}\right\} \! \in \! \mathbb{R}^{m \times D}$} \\ \bottomrule
\end{tabular}
}
\caption{The difference between APE and our feature selection method. There is functional overlap between inter-class similarity and inter-class variance in APE. Our method formulates rules from both intra-class and inter-class perspectives to achieve better performance.}
\label{tab:ape_diff}
\end{table}

\textbf{APE}.
There are two rules for feature selection of APE. 
\textbf{(1)} Minimizing inter-class similarity $\boldsymbol{S}_{inter}$: APE calculates the inter-class similarity between different text features (\textit{e.g.} $\boldsymbol{t}_{n_1}^i, \boldsymbol{t}_{n_2}^j$) of different categories (\textit{e.g.} $n_1, n_2$) at the channel level and minimizes it. 
\textbf{(2)} Maximizing inter-class variance $\boldsymbol{V}$: APE calculates the variance of average features ($\textit{e.g.} \{\boldsymbol{t}_1^{\top}, \boldsymbol{t}_2^{\top}, \ldots, \boldsymbol{t}_m^{\top}\}$) across different categories (\textit{e.g.} $1,2, \ldots, m$) at the channel level and maximizes it. However, we find that \textbf{these two rules are redundant, as they serve the same purpose}, which is to make the features between different categories as dissimilar as possible while ignoring the intra-class similarity between different features within the same class.

\textbf{Ours}.
We specify rules from both intra-class and inter-class perspectives. 
\textbf{(1)} Minimizing intra-class similarity $\boldsymbol{S}_{intra}$: calculating the similarity between different text features (\textit{e.g.} $\boldsymbol{t}_{n}^i, \boldsymbol{t}_{n}^j$) within the same category (\textit{e.g.} $n$) and minimizing it. The purpose of this rule is to \textbf{avoid redundancy between different text features and preserve the semantic diversity brought by text enrichment}. 
\textbf{(2)} Maximizing inter-class variance $\boldsymbol{V}$: This rule remains consistent with APE, aiming to maximize the semantic distance between different categories as much as possible.

We conduct a comparison in Tab.~\ref{tab:selection_comp}, and our improved method (case \#2) achieves a performance gain of 1.15\% compared to the original APE (case \#5).

\subsection{Text Enrichment}

Text descriptions are formulated through handcraft templates or generated through GPT-3. We obtained around 15 descriptions for each category.

The \textbf{handcraft template} of point cloud/object labels in segmentation and detection tasks is ``\texttt{The point cloud of a \{CLS}\}'', where CLS represents the category. As for the scene classification task, the handcraft template is ``\texttt{This is a 3D indoor room, and the room type is a \{CLS\}}'', where CLS denotes the specific room type.

\textbf{GPT-based text generation} is returned through a Q\&A style. Tab~\ref{tab:gpt} presents examples of dialogues.

\begin{table}[!htbp]
\vspace{-5pt}
\setlength{\abovecaptionskip}{0pt}
\begin{tcolorbox}%[colback=white,colframe=black]
\textcolor[rgb]{0.0,0.7,0}{\textbf{Input Instruction:}} \par
There are 20 semantic categories in indoor scenes as follows: \par
``wall, floor, cabinet, bed, chair, sofa, table, door, window, bookshelf, picture, counter, desk, curtain, refridgerator, shower curtain, toilet, sink, bathtub, other furniture."

Do you know how to distinguish them? You can consider from the following perspectives:\par
$\bullet$ Describe what a \{\} looks like. \par
$\bullet$ Visually describe a \{\}. \par
$\bullet$ How can you identify a \{\}? \par
$\bullet$ Provide a visual analysis of a \{\} and its key components. \par
$\bullet$ Can you provide a detailed description of the \{\}'s physical appearance? \par
$\bullet$ What are the distinguishing features of a \{\}? \par
$\bullet$ A caption of a photo of a \{\}.\par
Please generate 15 descriptions for the ``table".
\par

\tcblower % divider

\textcolor[rgb]{0,0.0,0.7}{\textbf{GPT Response:}}  \par
\#1 A table is a flat, elevated surface used for various activities like dining or working. \par
\#2 A typical table's size and shape vary, often rectangular, round, or square, with various dimensions. \par
$\dots \dots$

\end{tcolorbox}
\caption{Fine-grained description generation based on GPT. Top: the instruction we provide. Bottom: The answers generated by GPT.}
\label{tab:gpt}
\vspace{-15pt}
\end{table}

\subsection{Training Details}

The code is implemented based on PyTorch.

$\bullet$ \textbf{Semantic segmentation on ScanNet} datasets (20, 200, and 485): Training data includes point coordinates, colors, and normals. We utilize the AdamW optimizer with an initial learning rate of 0.005 and a weight decay of 0.02. OneCycleLR for learning rate scheduling. These above settings, alongside data augmentation, align with those of PointTransformerV2 (PTv2)~\cite{wu2022point}. Training for 900 epochs with a batch size of 12 on four 24-GB NVIDIA-RTX-4090 GPUs takes about 22 hours.

$\bullet$ \textbf{Semantic segmentation on the S3DIS} dataset:  Training data includes point coordinates and colors. We employ the AdamW optimizer with an initial learning rate of 0.006 and a weight decay 0.05. MultiStepLR for learning rate scheduling. These above settings, alongside data augmentation, align with those of PTv2. Training for 3000 epochs with a batch size of 12 on four 24-GB NVIDIA-RTX-3090 GPUs takes about 18 hours.

$\bullet$ \textbf{Object detection} on the ScanNet dataset: Training data consists of point coordinates (the impact of colors on performance seems relatively minor), with 50,000 points sampled per scene. The AdamW optimizer is employed with initial learning rates of 2e-3 and 2e-4, along with a weight decay 5e-4. MultiStepLR for learning rate scheduling. ScanNet485 is used for joint training. The above settings, along with data augmentation, align with BUTD-DETR~\cite{jain2022bottom} and EDA~\cite{wu2023eda}. Training for 300 epochs with a batch size of 12 on four 24-GB NVIDIA-RTX-3090 GPUs takes about 14 hours.

$\bullet$ \textbf{3D scene classification} on the ScanNet dataset: Hyperparameters such as learning rates are consistent with those used for segmentation tasks on ScanNet. Training for 600 epochs with a batch size of 12 on two 24-GB NVIDIA-RTX-3090 GPUs takes about 12 hours.

%%%%%%%%%%%%%%%%%%%%%%%%%%%%%%%%%%%%%%%%%%%%%%%%%%%
% 2. Additional experiments
%%%%%%%%%%%%%%%%%%%%%%%%%%%%%%%%%%%%%%%%%%%%%%%%%%%
\section{Additional Experiments}
\label{sec:supp_exp}

~~~~\textbf{(1) Influence of Text Enrichment and Feature Selection on Semantic Concept.}

\begin{figure*}[htbp]
\centering
\includegraphics[width=1.0 \textwidth]{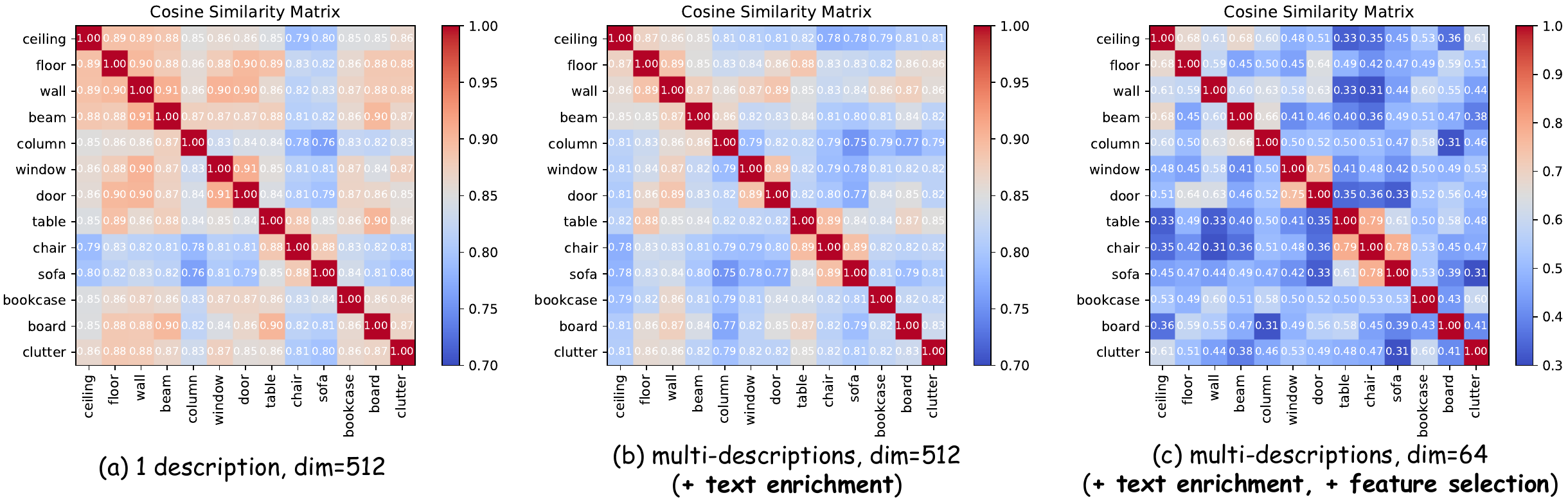} 
\caption{
Cosine similarity among textual features of the 13 categories within the S3DIS dataset. After text enrichment and feature selection, feature distinctiveness is enhanced while preserving the original semantic correlations.
}
\label{fig:sim_matrix}
%\vspace{-15pt}
\end{figure*}

To demonstrate the enhancement of feature distinctiveness through text enrichment and feature selection, we compute cosine similarity on the textual features of 13 categories within the S3DIS dataset. 
\textbf{\romannumeral1)} In Fig.~\ref{fig:sim_matrix} (a), only one text description is constructed for each category, and we extract their original 512-dimensional features. It can be observed that aside from the highest similarity with itself, different category features exhibit significant similarity, which blurs the training of distinctive features. 
\textbf{\romannumeral2)} As depicted in Fig.~\ref{fig:sim_matrix} (b), text enrichment is achieved through LLM-based text generation. After computing the average feature of multiple descriptions for the same category, the similarity matrix demonstrates improved distinctiveness, illustrating that fine-grained descriptions are advantageous for comprehending semantic concepts. 
\textbf{\romannumeral3)} Fig.~\ref{fig:sim_matrix} (c) illustrates the results after feature selection, where feature distinctiveness is further enhanced without compromising the original semantics. For instance, high semantic correlations remain between ``ceiling'' and ``floor'', ``window'' and ``door'', ``chair'' and ``sofa''.

% tab: similarity matrix

\textbf{(2) Compared with Other Feature Selection Methods.} 

In the main text, we demonstrate the advantages of our proposed statistical-based significance feature selection method over retaining the original features and trainable projection layers. Here, we compare with other feature selection methods: 

$\bullet$ Randomly selecting 64 dimensions from the original 512 dimensions. (case \#3) \par
$\bullet$ Average pooling the 512 dimensions to 64 dimensions.(case \#4) \par
$\bullet$ The original APE~\cite{zhu2023not} selects 64 dimensions from the 512 dimensions through the two criteria discussed in Sec.~\ref{sec:com_ape}. (case \#5) 

The results are shown in Tab.~\ref{tab:selection_comp}. \textbf{\romannumeral1)} Our method (case \#2) still outperforms case \#3, \#4, and \#5, indicating the effectiveness of the proposed approach. \textbf{\romannumeral2)} Random selection (case \#3) even achieves higher performance than keeping the original features (case \#1), indicating the existence of information redundancy in high-dimensional features. \textbf{\romannumeral3)} The pooling method (case \#4) not only disrupts the original features but also diminishes performance.
\textbf{\romannumeral4)} Our method achieves a 1.15\% performance improvement compared to the original APE (case \#5), demonstrating the validity of the two rules for intra-class and inter-class that we discussed in Sec.~\ref{sec:com_ape}.

\begin{table}[t]
\small
\centering
\setlength{\tabcolsep}{15pt}
\begin{tabular}{cc|c}
\toprule
Case    & Method      & mIoU  \\ \midrule
\#1 & 512 (Orig.) & 30.37 \\
\textbf{\#2} & \textbf{64 (Select, Ours)} & \textbf{33.29} \\ \midrule
\#3 & 64 (Random) & 31.18 \\
\#4 & 64 (Pool)   & 29.96 \\ 
\#5 & 64 (Select, APE) & 32.14 \\
\bottomrule
\end{tabular}
\caption{Comparison of feature selection methods on ScanNet200. Case \#1: Keeping the original 512-dimensional features. Case \#2: Our proposed statistical-based significance feature selection. Case \#3: Randomly selecting 64 dimensions. Case \#4: Average pooling to 64 dimensions. Case \#5: The original APE feature selection criteria.}
\label{tab:selection_comp}
\end{table}

\textbf{(3) Ablation on Retained Channel Number.}

To find the trade-off between performance and computation regarding the decision of how many channels to preserve during feature selection, we experimented with keeping 32, 64, and 128 channels, respectively. The results are presented in Tab.~\ref{tab:chanel} Overall, all three choices outperform retaining the original features (case \#1), underscoring the redundancy and ambiguity in high-dimensional features. Retaining 32 channels may lead to excessive feature compression, losing information and the poorest performance. In our implementation, selecting 64 channels achieved optimal performance while significantly reducing computational load.

\begin{table}[t]
\small
\centering
\setlength{\tabcolsep}{15pt}
\begin{tabular}{cc|c}
\toprule
 Case   & Method       & mIoU  \\ \midrule
\#1 & 512 (Orig.)  & 30.37 \\ \midrule
\#2 & 32 (Select)  & 30.63 \\
\textbf{\#3} & \textbf{64 (Select)}  & \textbf{33.29} \\
\#4 & 128 (Select) & 31.62 \\ \bottomrule
\end{tabular}
\caption{Ablation study on retaining different numbers of channels for 512-dimensional Textual Features. In our implementation, we select 64 channels. Evaluated on ScanNet200.}
\label{tab:chanel}
\end{table}

\begin{table}[h]
\small
\centering
\setlength{\tabcolsep}{3pt}
\resizebox{\columnwidth}{!}{
\begin{tabular}{ccc|ccc}
\toprule
\multicolumn{3}{c|}{Training}            & \multicolumn{3}{c}{Inference}             \\
\scriptsize{ScanNet20} & \scriptsize{ScanNet200} & \scriptsize{ScanNet485} & \scriptsize{ScanNet20} & \scriptsize{ScanNet200} & \scriptsize{ScanNet485} \\ \midrule
\ding{51}         & \ding{51}          &    -          & 75.1     & 29.5       & -          \\
    -      & \ding{51}          & \ding{51}            & -         & 32.9      & 13.8      \\ \bottomrule
\end{tabular}
}
\caption{Our method enables joint training on multiple datasets and allows a unified model to achieve inference across multiple benchmarks. Measured by mIoU.}
\label{tab:joint_traing}
%\vspace{-5pt}
\end{table}

\textbf{(4) The Flexibility and Robustness of Text Queries.}
1) Flexibility in training. As analyzed in Sec.~\ref{sec:analysis}, we can jointly train multiple benchmarks with different numbers of categories, as shown in Tab.~\ref{tab:joint_traing}. We only need to ensure that the dimensions of the textual features are consistent with those of the point cloud features.
2) Flexibility in inference. During inference, we use the offline-preserved average features of 15 text descriptions of each category. These average features serve as supervisory signals during training. To prove that the model learns meaningful semantic concepts rather than fixed average text from training, as depicted in Tab.~\ref{tab:text_query}, we utilize three LLMs to generate one description for each category in ScanNet20 randomly. Due to the diversity of models and the stochastic nature of the generation process, these \textbf{texts are nearly unseen during training} (but exhibit consistent semantics). The results indicate that texts from different models achieve comparable performance, validating the model learned meaningful semantic concepts and possessing flexibility and robustness of text.

\begin{table}[t]
\small
\centering
\setlength{\tabcolsep}{10pt}
\begin{tabular}{cc|ccc}
\toprule
\multicolumn{2}{c|}{LLM}      & mIoU  & mAcc  & allAcc    \\ \midrule
\#1  & GPT-4        & 71.3 & 83.3 & 89.0    \\
\#2  & Claude         & 70.9 & 81.4 & 88.5 \\
\#3  & Bard           & 70.6 & 81.2 & 88.1 \\ \bottomrule
\end{tabular}
\caption{Three different LLMs randomly generated one text description for each category of ScanNet20, and these texts are nearly unseen during training. The comparable performance indicates that the model acquires meaningful semantic concepts rather than relying on the fixed text that appears during training. GPT-4 \url{https://openai.com/gpt-4}. Claude \url{https://claude.ai/}. Bard \url{https://bard.google.com/}.}
\label{tab:text_query}
\end{table}

\begin{table}[t]
\small
\centering
\setlength{\tabcolsep}{4pt}
\resizebox{\columnwidth}{!}{
\begin{tabular}{c|ccc|c}
\toprule
Method        & Head          & Common        & Tail          & \textbf{All}           \\ \midrule
\scriptsize{Mink.Net ~\cite{choy20194d}}    & 48.3          & 19.1          & 7.9           & 25.1          \\
SupCon\scriptsize{~\cite{khosla2020supervised}}        & 48.5          & 19.1          & 10.3          & 26.0            \\
CSC\scriptsize{~\cite{hou2021exploring}}           & 49.4          & 19.5          & 10.3          & 26.5          \\
LG\scriptsize{~\cite{rozenberszki2022language}}            & 51.5          & 22.7          & 12.5          & 28.9          \\
CeCo\scriptsize{~\cite{zhong2023understanding}}          & 51.2          & 22.9          & 17.1          & 30.3          \\
CeCo (lovász)~$\star$\scriptsize{~\cite{zhong2023understanding}}  & 52.4 & 26.2          & \textbf{17.9}          & 32.0            \\
PTv2\scriptsize{~\cite{wu2022point}}          & 53.0          & 27.1          & 15.3          & 31.9          \\ \midrule
\textbf{LAST-PCL (Ours)}          & \textbf{53.9}          & \textbf{29.0} & 17.1 & \textbf{33.3} \\ \bottomrule
\end{tabular}
}
\caption{Performance on the three subsets of ScanNet200 \textbf{validation set}.
Three subsets comprise 66, 68, and 66 categories, respectively. $\star$ indicates the version using the semantic segmentation-specific loss (Lovász loss).}
\label{tab:scannet200_subset}
\end{table}

\begin{table}[]
\small
\centering
\resizebox{\columnwidth}{!}{
\renewcommand{\arraystretch}{1.1}
\begin{tabular}{c|ccc|c}
\toprule
Method          & Head & Common         & Tail & \textbf{All}  \\ \midrule
Mink.Net      & 46.3 & 15.4          & 10.2 & 25.3 \\
CSC             & 45.5 & 17.1          & 7.9  & 24.9 \\
LG              & 48.5 & 18.4          & 10.6 & 27.2 \\
OctFormer\scriptsize{~\cite{Wang-2023-octformer}}       & ~~53.9$^2$ & ~~26.5$^2$          & 13.1 & ~~32.6$^3$ \\
CeCo (lovász)   & 52.1 & 23.6          & ~~15.2$^3$ & 31.7 \\
CeCo$\ddagger$(lovász)   & ~~55.1$^1$ & ~~24.7$^3$          & ~~18.1$^1$ & ~~34.0$^1$   \\ \midrule
\textbf{LAST-PCL (Ours)} & ~~53.3$^3$ & ~~27.9$^1$ & ~~15.5$^2$ & ~~33.6$^2$ \\ \bottomrule
\end{tabular}
}
\caption{Performance on ScanNet200 \textbf{test set}. $\ddagger$ denotes the enhanced version of the ensemble of three methods. All numbers are from the online benchmark on August 2023.}
\label{tab:scannet200_test}
\end{table}

\balance

\textbf{(5) Detailed Results on ScanNet200.}

The ScanNet200 dataset divides categories into three subsets based on semantic instance frequency (head-shot, medium-shot, and few-shot): \textbf{\textit{head}}, \textbf{\textit{common}}, and \textbf{\textit{tail}}. 
\textbf{\romannumeral1)} We present a comprehensive comparison with SOTA in Tab.~\ref{tab:scannet200_subset}.
LAST-PCL achieves top-2 performance across all three subsets, only slightly trailing behind in the ``tail'' subset. 
LAST-PCL substantially improves over the baseline (PTV2), demonstrating the benefits of introducing out-of-domain prior knowledge through text contrastive training.
\textbf{\romannumeral2)} Additionally, we present the results of the ScanNet200 test set in Tab~\ref{tab:scannet200_test}. We achieve second-place performance overall and secure the first rank in the ``common'' subset. The evaluation of the validation and test sets validates LAST-PCL as a promising approach.

\vspace{10pt}
%%%%%%%%%%%%%%%%%%%%%%%%%%%%%%%%%%%%%%%%%%%%%%%%%%%
% 3. Visualization Results
%%%%%%%%%%%%%%%%%%%%%%%%%%%%%%%%%%%%%%%%%%%%%%%%%%%
\section{Visualization Results}
\label{sec:supp_vis}
\vspace{10pt}

\begin{figure*}[!h]
\centering
\setlength{\abovecaptionskip}{0pt}
\includegraphics[width=1.0 \textwidth]{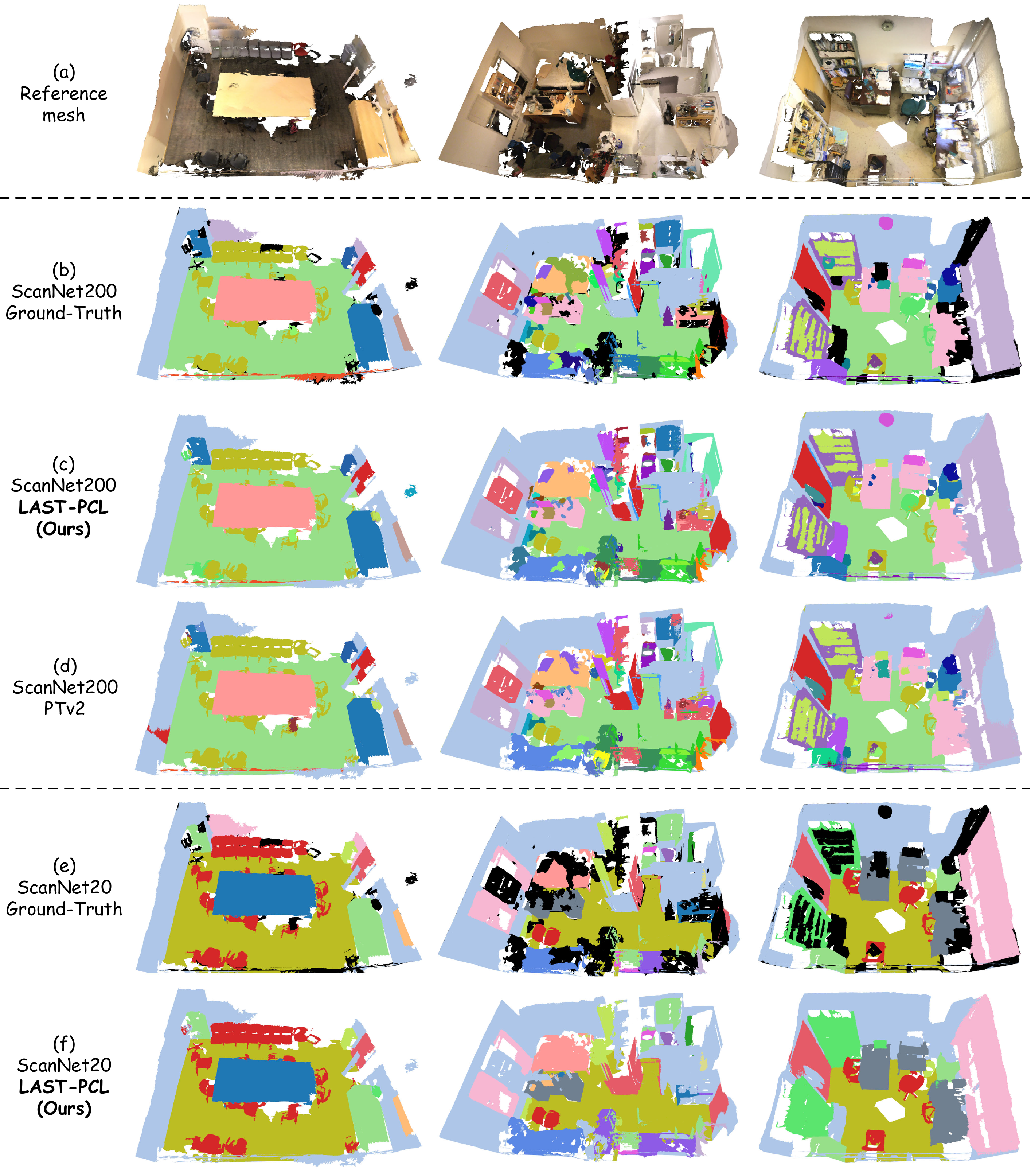} 
\caption{
Semantic segmentation visualization on ScanNet200 and ScanNet20. (a) Reference mesh. (b-d) Ground-truth and predicted results on ScanNet200. (e-f) Ground-truth and predicted results on ScanNet20. Note that the black regions in (b,e) represent unannotated points.
}
\label{fig:vis}
%\vspace{-15pt}
\end{figure*}

~~\textbf{Semantic Segmentation Visualization.} We present the results on ScanNet200 and ScanNet20 in three scans in Fig.~\ref{fig:vis}. LAST-PCL achieves outstanding segmentation performance. Compared to PointTransformerV2 (PTv2), LAST-PCL exhibits more robust semantic recognition capabilities at a finer granularity.

\begin{figure*}[t]
\centering
\setlength{\abovecaptionskip}{0pt}
\includegraphics[width=1.0 \textwidth]{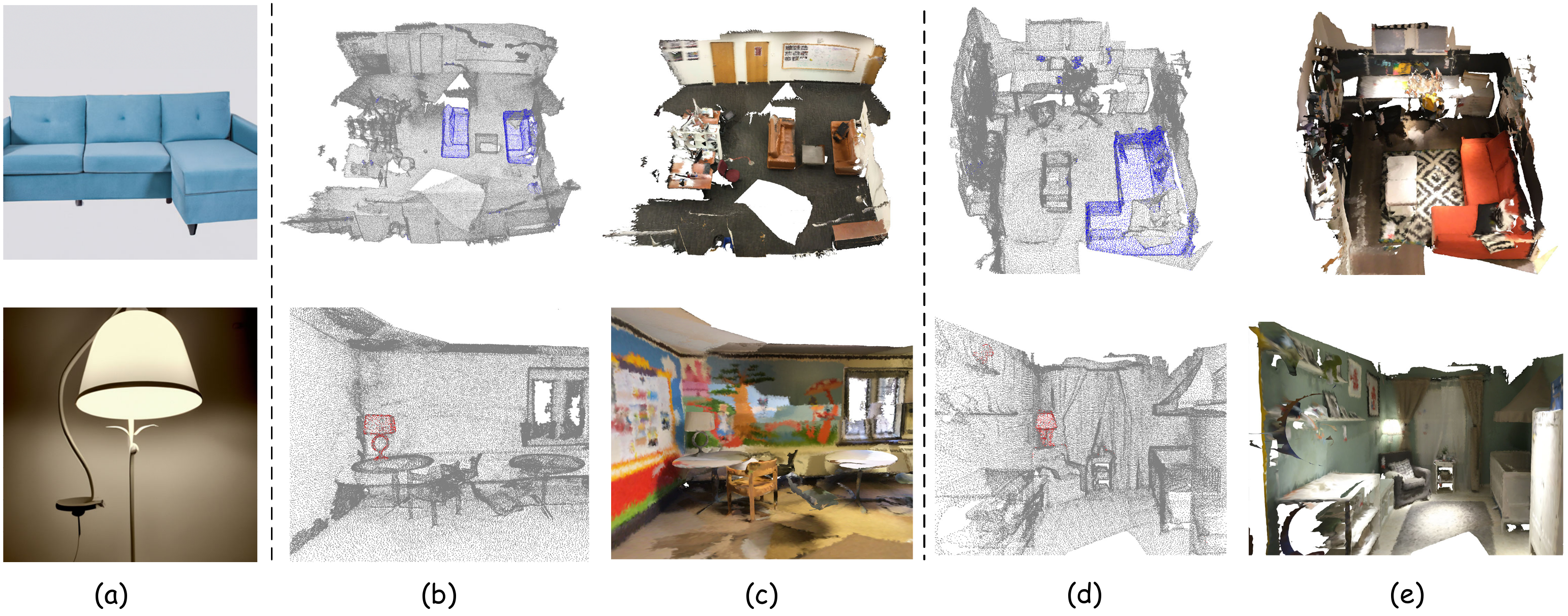} 
\caption{
Image query visualization. (a) Query images generated by DALL·E. (b, d) Point Cloud. (c, e) Reference mesh.
}
\label{fig:img_query}
%\vspace{-15pt}
\end{figure*}

\textbf{Image Query Visualization.} The results for image queries are shown in Fig.~\ref{fig:img_query}. Consistent with the setup in the main text, we employ DALL·E~\cite{ramesh2021zero} to generate images (Fig.~\ref{fig:img_query} (a)), extract 512-dimensional image features using CLIP's image encoder, and then retrieve the top-64 features using our offline preserved channel ranking. These features are used to compute similarity scores with point features. The colored regions in (Fig.~\ref{fig:img_query} (b, d)) indicate point clouds with high similarity to the query image. The version of the pre-trained CLIP model we used is ViT-B/32.

\begin{figure*}[t]
\centering
\setlength{\abovecaptionskip}{0pt}
\includegraphics[width=1.0 \textwidth]{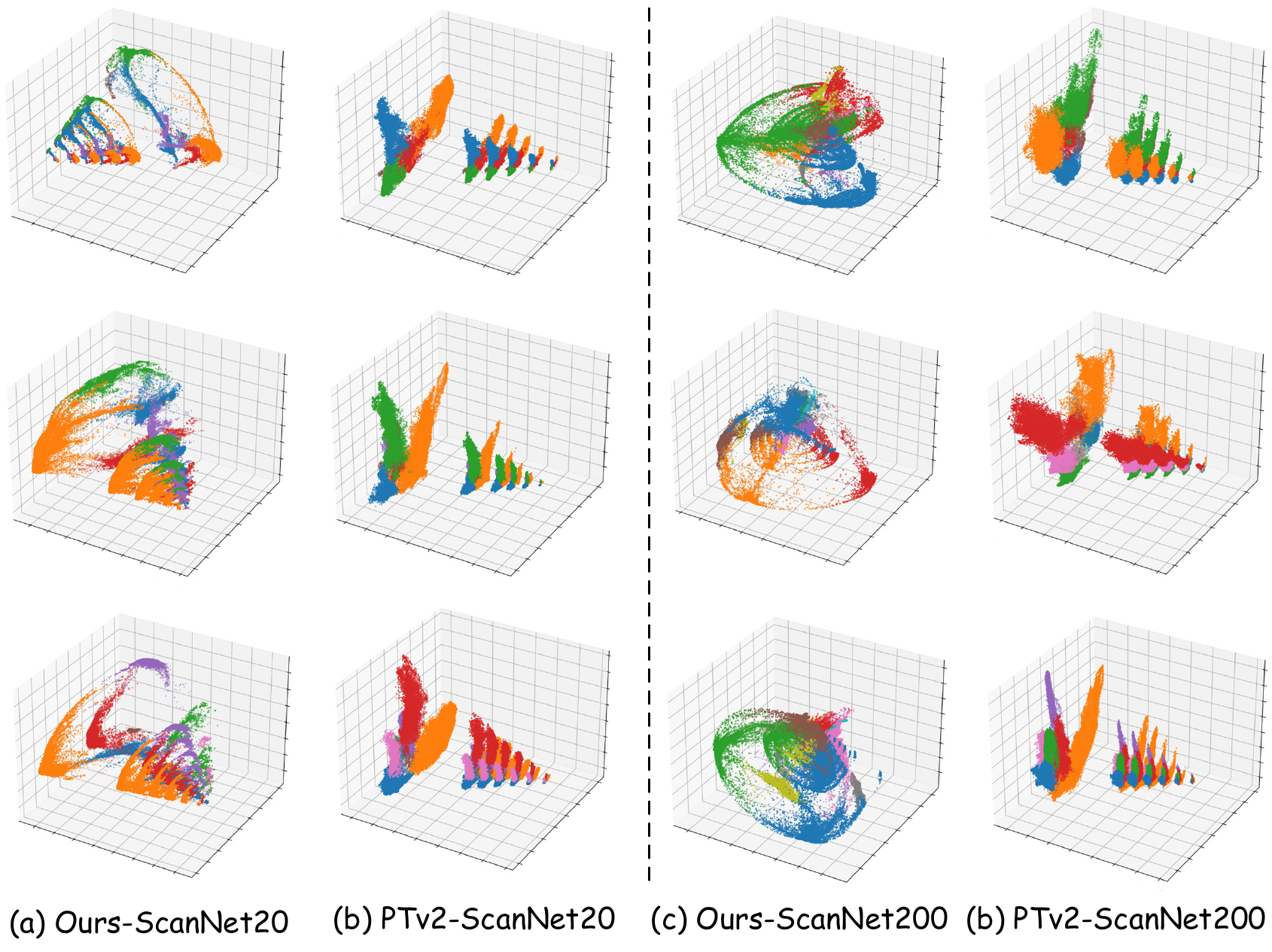} 
\caption{
Feature space visualization. (a, c) Features learned by LAST-PCL exhibit continuity, reflecting semantic correlations. (b, d) Features learned by PTv2 based on one-hot label supervision show discreteness.
}
\label{fig:supp_feat_vis}
%\vspace{-15pt}
\end{figure*}

\textbf{Feature Space Visualization.} In Fig.~\ref{fig:supp_feat_vis}, we provide a comparison of feature distributions across multiple samples. Consistent with the conclusions drawn in the main text, Features trained through text contrastive learning exhibit continuity, facilitating the learning of continuous and semantically meaningful concepts.

